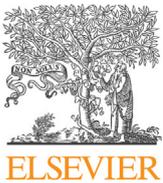
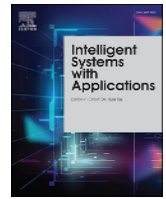

Review

# Continual learning for predictive maintenance: Overview and challenges

Julio Hurtado [a,*], Dario Salvati [a], Rudy Semola [a], Mattia Bosio [b,c], Vincenzo Lomonaco [a]

[a] *Department of Informatics, University of Pisa, Largo Bruno Pontecorvo, 3, Pisa, 56127, Italy*
[b] *SEA Vision, via Claudio Treves, 9E, Pavia, 27100, Italy*
[c] *ARGO Vision, via Gianfranco Zuretti, 4, Milano, 20125, Italy*

A B S T R A C T

Deep learning techniques have become one of the main propellers for solving engineering problems effectively and efficiently. For instance, Predictive Maintenance methods have been used to improve predictions of when maintenance is needed on different machines and operative contexts. However, deep learning methods are not without limitations, as these models are normally trained on a fixed distribution that only reflects the current state of the problem. Due to internal or external factors, the state of the problem can change, and the performance decreases due to the lack of generalization and adaptation. Contrary to this stationary training set, real-world applications change their environments constantly, creating the need to constantly adapt the model to evolving scenarios. To aid in this endeavor, Continual Learning methods propose ways to constantly adapt prediction models and incorporate new knowledge after deployment. Despite the advantages of these techniques, there are still challenges to applying them to real-world problems. In this work, we present a brief introduction to predictive maintenance, non-stationary environments, and continual learning, together with an extensive review of the current state of applying continual learning in real-world applications and specifically in predictive maintenance. We then discuss the current challenges of both predictive maintenance and continual learning, proposing future directions at the intersection of both areas. Finally, we propose a novel way to create benchmarks that favor the application of continuous learning methods in more realistic environments, giving specific examples of predictive maintenance.

## 1. Introduction

In the last decade, electromechanical devices have been fundamental to industrial, scientific, and human development. Machines today are widespread, from assembly lines to hard disk drives, leading to various components, models, and upgrades within each application. Despite these improvements, devices are not immune to (hardware or software) issues, and monitoring and maintenance are necessary to ensure they continue working correctly.

The area of Predictive Maintenance (PdM) has proposed various techniques that have been implemented in a wide variety of scenarios and problems to carry out (semi) automated maintenance optimally, effectively, and preemptively. Following the big data and deep learning wave, a popular approach to facing this problem is to train models with previously collected data to infer when maintenance is necessary. These approaches are called data-driven strategies and have led to the continual improvement and development of the area, demonstrating encouraging results in the most popular benchmarks in PdM (Zonta et al., 2020, Serradilla et al., 2022).

However, a problem with deep learning solutions is that they are trained in a static scenario. This assumes that the data distribution used during training will not change and always reflect the whole set of observations the model will encounter. As we will argue in the following sections, this is not always the case. In real-world environments, the distribution of inputs and outputs may fluctuate over time following processes known as distribution drifts. By changing the distribution of the input or output data, the model will perform erratically, since it will receive unusual information from the input, mispredicting it and failing sooner than expected, and causing unexpected maintenance delays. Despite the progress made, such changes in distributions are still an open problem in the deep learning community (Hu et al., 2020).

An effective way to mitigate the drift, is to keep the models up-to-date with the current distribution. The area that studies these ideas are known as Continual Learning (CL) (Vijayan & Sridhar, 2021), and it aims to train models to adapt and continuously acquire updated knowledge over time. CL deals with scenarios in which the model must be trained sequentially on a set of tasks while retaining knowledge from previous tasks. Most CL methods focus on preventing knowledge loss of old distributions. This problem is known as catastrophic forgetting (McCloskey & Cohen, 1989), and is caused mainly by the change in the value of the model weights.

In general, the problem with most CL methods is that they focus mainly on synthetic benchmarks, often limited to random splits of known datasets. These benchmarks significantly differ from unstructured natural environments (Cossu et al., 2022a). Because many of these benchmarks are unrealistic, practical applications of CL methods have been limited to very specific cases.

This study has two main objectives. The first objective is to augment PdM with CL, by examining how non-stationary environments can affect PdM, and how CL can mitigate these foundational issues. The second objective is to boost CL development through PdM. Here we present why these non-stationary industrial scenarios can help to assess one of its limitations, the low portability in real-world after-deployment application contexts (Liu, 2020).

In line with the objectives, this work contributes as follows: (1) a brief review of the areas involved. Since we are proposing an intersection between different areas, we believe it is necessary for counterpart researchers to understand the current state of the areas involved. (2) A motivation for the integration of different approaches, proposing the limitations of established areas and how we can complement one's weaknesses with others' strengths. (3) After the integration, we present works that have contributed to this integration, either at the operational or methodological level. (4) Based on the current state, we discuss various limitations of current methods and suggest future research directions to expand these fields. (5) Guided by our main objective, we present a proposal for how future benchmarks in the area should be constructed, considering the challenges of PdM and the limitations of current CL benchmarks.

This work brings together different concepts and ideas. In Section 2, we will briefly explain predictive maintenance. Then, we will explore public datasets often used in PdM, evaluating their advantages and shortcomings. These drawbacks will open the discussion to non-stationary environments in Section 3, and how this can affect PdM models in real-world applications. Next, we will cover Continual Learning scenarios and methods in Section 4. Section 5 present current approaches that apply CL to real-world environments. Section 6 focuses on CL for PdM, current methods, challenges, and future directions. Encouraged by the most significant limitations to applying CL to PdM problems, in Section 7, we present the limitations of existing benchmarks and a research direction in detail.

## 2. Predictive maintenance

During the first years of the industrial revolution, maintenance was primarily reactive, which means machines were fixed only when inoperative. This maintenance strategy strongly affects productivity, since the machine becomes inactive for the time it takes to repair it. To solve this issue, an approach known as Preventive Maintenance (PM) (Ran et al., 2019) was proposed, involving periodic maintenance of machines and components to prevent long periods of inefficiency. PM solves the problem of preventative damage but has the inconvenience that maintenance is carried out without considering the status of the machines.

Significant technological advances, especially in sensors and the Internet of Things (IoT), have enabled machines to constantly monitor their operation, detecting productivity, efficiency, or effectiveness drops. These advancements have made it possible to have a large amount of data associated with machines and component status, allowing the expansion of artificial intelligence models to predict failures. This area is known as Predictive Maintenance (PdM) (Hansen et al., 1994), and aims to identify the right time to perform maintenance on a particular machine. This approach prevents severe failure that can leave a machine inoperative and, at the same time, decreases unnecessary maintenance or downtime.

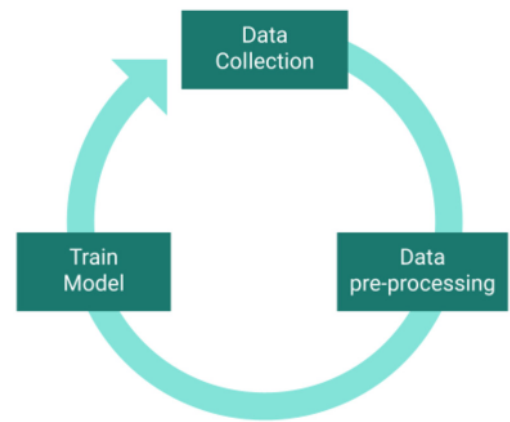

**Fig. 1.** Finding solutions based on ML is an iterative process consisting of data collection, pre-processing, and model training steps.

A number of factors play a role in determining when maintenance is needed. Broadly speaking, and based on Zonta et al. (2020), the techniques that deal with PdM can be divided into three groups: the first is referred to as physical model-based, which uses physics to predict the future behavior of the machine's components, and thus determines when it is about to fail. The second group is called knowledge-based methods and focuses on reducing the need for complex physical models by supplementing them with heuristics associated with each machine. Finally, the data-driven group consists of those methods that derive their predictions from extracting statistical patterns from the data, primarily through Machine Learning (ML) techniques (Ran et al., 2019, Zonta et al., 2020, Serradilla et al., 2022). Given the rapid growth of the latter and that our objective is to study continual learning as applied to this context, we will focus on Data-driven solutions in this survey.

Similar to most ML methods, for a data-driven solution, one can identify three central components (Van Tung & Yang, 2009), shown in Fig. 1. The first component is data acquisition, which consists of gathering extensive data related to the given problem. The second component is pre-processing the data. Due to the naturalness of the data, it may be necessary to clean incomplete data or remove outliers. Finally, the third component is the selection, implementation, and training method. It is imperative to note that this is an iterative process, where the additional data collected can increase the diversity of the training set (Diallo et al., 2021).

It is worth clarifying that this article is not intended to be a complete survey for the PdM task. To better understand this topic, we invite the reader to consult the following works: in Ran et al. (2019), the authors present a complete overview of the system architecture, purposes, and approaches for PdM. Then, Zonta et al. (2020) updates the literature review and introduces a new taxonomy to classify the different methods of this research area. Finally, Serradilla et al. (2022) focused their survey on Deep Learning (DL) methods, comparing different architectures on a known dataset.

### 2.1. Problems

Generally speaking, PdM aims to reduce operational expenses from two angles. The first is minimizing the time an asset spends in maintenance due to a failure detected too late. The second aims to limit maintenance performed on a machine, performing it only when necessary.

Concerning the first angle, fault diagnosis helps recognize failures as soon as possible, isolating and identifying faulty components. By detecting that something is wrong compared to average performance, maintenance can be carried out before it becomes critical. Although most works focus on detecting failures, some also concentrate on the location and identification of the components causing the problem and identifying the leading cause of the fault (Liu et al., 2018b).

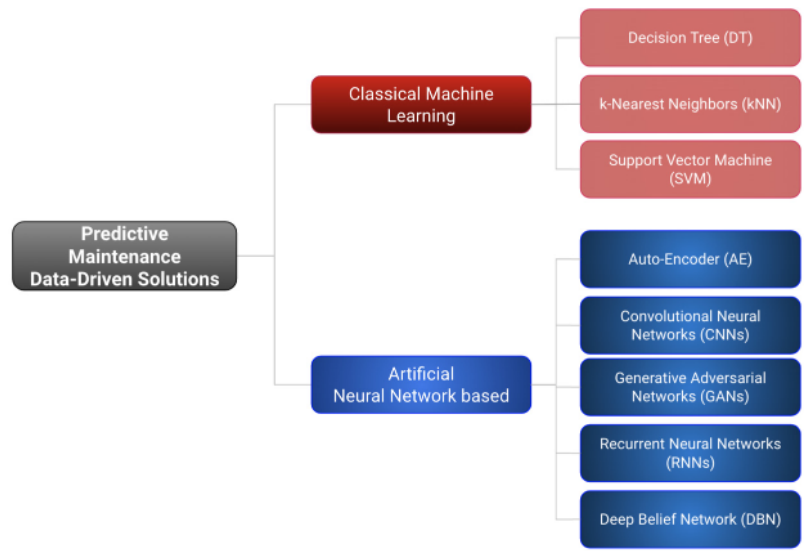

**Fig. 2.** Graphical representation of the division of PdM data-driven solutions.

Concerning the second angle, to minimize the amount of maintenance done to an asset, one should estimate how much time is left until the machine fails. Two states of operation can be identified in an asset using this observation: 1) healthy and 2) unhealthy (Lei et al., 2018). fault prognosis is applied in this second stage to predict the Remaining Useful Life (RUL) the asset has left before requiring maintenance. With fault prognosis, it is possible to schedule maintenance better and minimize total maintenance (Rezamand et al., 2020, Gao & Liu, 2021).

### 2.2. Approaches

The outbreak of ML has invited the PdM community to use different data-driven solutions to detect future failures (Baltazar et al., 2018, Zhang et al., 2020b). The advantage of ML-based techniques is that they can use past experiences to find patterns that identify current defects and then use those patterns to generalize to novel data.

Despite having different objectives, fault diagnosis and fault prognosis use the same base models to obtain their predictions. We can divide these base models into classical ML and artificial neural network approaches, as shown in Fig. 2. As opposed to the former, the latter learns representations alongside the classifier, rather than requiring a manually created feature vector. The following is a partial list of models used in PdM.

#### 2.2.1. Classical machine learning approaches

**Decision Trees**: DTs are non-parametric methods that learn decision rules from input data. Starting from a root node, the model makes decisions that allow it to go through different paths until it reaches a leaf node representing a prediction. DT techniques have been frequently utilized for fault diagnosis in numerous scenarios, such as refrigerant flow system (Benkercha & Moulahoum, 2018), anti-friction bearing (Patil & Phalle, 2019), and grid-connected photovoltaic system (Benkercha & Moulahoum, 2018). The speed of training of these models, plus the rapid prediction, have helped us to use them in real-time problems (Kou et al., 2019, Ayvaz & Alpay, 2021). The simplicity and variety of DT-based methods allow them to be used with positive results in fault prognosis scenarios, such as wind turbine engine (Mathew et al., 2017), lithium-ion battery (Zheng et al., 2019), and mechanical equipment (Bakir et al., 2019).

**k-Nearest Neighbors**: kNN is an algorithm that does not require training. When an instance arrives, the distance between this element and all the training set elements is computed to find the $k$ instances closest to it. The class is assigned based on majority voting between those $k$ instances. Multiple proposals that use kNN have been presented for fault diagnosis. Some use it in combination with PCA (Tian et al., 2015), hierarchical methods (Baraldi et al., 2016), or distance and similarity density (Uddin et al., 2016, Appana et al., 2017). Other applications of kNN include photovoltaic system faults (Madeti & Singh, 2018) and petrochemical rotating faults (Xiong et al., 2015). Since RUL is a continuous feature problem, kNN is challenging. Even so, Liu et al. (2017d) applies it to power plants, in conjunction with evidential theory, to predict degradation.

**Support Vector Machine**: A SVM is a supervised method that aims to find a hyperplane that optimally divides two classes. To generate this hyperplane, SVM maximizes the margin between the Support Vectors (the elements closest to this hyperplane). As a result of the positive results achieved with SVMs, they have been applied to many different problems. These include wind turbines (Santos et al., 2015), bearing (Soualhi et al., 2014), chillers (Han et al., 2019), and rotation machinery (Zhu et al., 2018). Similar to SVM, but for regression problems, Support Vector Regression (SVR) searches for the hyperplane containing the most points. This approach has been used in various RUL problems, such as lithium-ion battery (Wei et al., 2017, Zhao et al., 2018, Du et al., 2018) and bearings (Sui et al., 2019).

#### 2.2.2. Artificial neural networks based approaches

**Artificial Neural Networks**: an ANN consists of an input layer, with one or more hidden layers, followed by an output layer. The output of each layer passes through a (usually nonlinear) activation function. Given the versatility of these architectures, they can be applied both to detection (Samanta & Al-Balushi, 2003) and identification of faults (Sreejith et al., 2008). These approaches can also be used for fault prognosis. Teng et al. (2016), Elforjani and Shanbr (2017) used an ANN to predict the tendency of the feature series for wind turbine degradation.

**Auto-Encoder**: The main objective of AEs is to learn a high-level representation from the raw data. An Encoder is trained to compress the representation of an input $x$, and then a second model, the Decoder, decompresses the vector to generate the same input $x$. Previous works used AEs to capture valuable features for fault diagnosis (Shao et al., 2017b, Haidong et al., 2018) and fault prognosis (Xia et al., 2018, Ma et al., 2018) problems. The representations found by the AEs can be used in conjunction with other models, such as ANNs (Sun et al., 2016), or directly for detecting anomalies (Chao et al., 2021).

**Convolutional Neural Networks**: CNNs can find local patterns via convolutional operations, which are combined to generate global representations of the input. It has been proven that finding local patterns works well in structured data, like images (Wang et al., 2019b, Oh & Jeong, 2019, Jia et al., 2019, Liu et al., 2017e), but can likewise be applied to 1D inputs (Chen et al., 2015, Kiranyaz et al., 2018, Li et al., 2019b, Ince et al., 2016). CNN can also be applied to handle heterogeneous data, as shown by Mezair et al. (2022). Some studies have used time frames to predict the degradation in time of an asset (Sateesh Babu et al., 2016, Yang et al., 2019). Furthermore, CNN was also used with a joint-loss function between fault recognition and RUL (Liu et al., 2019).

**Recurrent Neural Networks**: Machines fail over time, so it is natural to study the effect of data sequences to predict their failure. RNNs work with a stream of data, where the current frame output determines the future representation together with the next frame. Previous works have applied this technique to different fault diagnosis scenarios, such as rotating machinery (Li et al., 2018, Yang et al., 2018), chemical processes (Yuan & Tian, 2019), or rolling bearing (Zhao & Shao, 2020). For fault prognosis, LSTMs (a variation of classical RNNs) have been used on problems of voltage prediction (Hong et al., 2019) and aircraft engines (Wu et al., 2018, 2020). In addition, RNNs have also been used with multiple losses to predict degradation assessment and RUL (Miao et al., 2019).

**Deep Belief Network**: Similar to deep neural networks, DBN stacks multiple layers to extract high-level features from the data. The difference is that each layer is constructed with Restricted Boltzmann Machines. Similarly to previous architectures, DBN can be applied to various fault diagnosis problems, such as axial piston (Wang et al., 2018), rolling bearings (Shao et al., 2017a), among others (Tamilselvan & Wang, 2013). DBN was also used with PCA (Zhu & Hu, 2019), or AE (Chen & Li, 2017). Concerning fault prognosis, DBN has been used in both wind turbines (Wang et al., 2019a) and lithium-ion batteries (Zhao et al., 2017).

**Generative Adversarial Networks**: GANs consist of two models, the first being a *generator* that generates data to trick a *discriminator*. Both are trained in parallel to improve the whole model's performance. In PdM, one problem is the imbalance of classes since it is easier to get data from machines working correctly than from failures. For this reason, GANs are mainly used to generate data that follow a similar distribution to the training set (Ding et al., 2019, Mao et al., 2019). GANs can also be used to detect anomalies (Akcay et al., 2018). An alternative use was proposed by Khan et al. (2018), Zhai et al. (2021), where they estimate future trajectories of the machine health indicators.

*2.2.3. Limitations*

Despite the different solutions presented, the proposed methods still have limitations. Firstly, the limited amount of training data restricts its applicability to closed and controlled environments. Many works use or generate synthetic datasets hoping to extrapolate to actual data, which is not always feasible. On the other hand, when access to large data is possible, it can be noisy or unbalanced by the very nature of the problem. A second limitation of these solutions is the lack of adaptability to evolving scenarios. A common solution is to completely retrain the model with old and new data each time updated data arrives. However, this becomes infeasible when data increases exponentially.

*2.3. Benchmarks*

Deep Learning's success in the last few years is primarily attributed to the large number of datasets available to train different kinds of models (Devlin et al., 2018, Bommasani et al., 2021, Radford et al., 2021, Reed et al., 2022). Multiple public datasets have emerged to help expand the area to various problems such as natural language processing, computer vision, and robotics. This approach encourages healthy competition to improve methods and strategies.

On the other hand, PdM lacks such datasets and benchmarks. Most publications consider a single benchmark in their experiments, and it is usually unavailable to the public. As research in this area grows, it becomes necessary to use a variety of benchmarks to verify the process of different methods. Ideally, these benchmarks should not only be public, but they must also meet multiple requirements to be a valid comparison between methods (Diallo et al., 2021, Hagmeyer et al., 2021).

Some public datasets have been proposed in the last few years, helping expand PdM, though not all meet the requirements to be a suitable benchmark. Table 1 contains a non-exhaustive list of datasets. Additionally, other datasets may be used indirectly for PdM, such as when a product is defective on a production line (Bosch, 2016) or in a chemical process (Chen, 2019). Current benchmarks have proven helpful to the community, but there is a tight coupling between the solution and the problem it aims to tackle. Due to issues such as limited data, class imbalance, and missing data, better benchmarks are needed. Possible solutions are simulation datasets (Zhang et al., 2019b); however, models trained in these datasets tend to lack generalization capabilities to other distributions.

According to Diallo et al. (2021), two datasets meet the requirements to be suitable benchmarks: TurboFan (Saxena & Goebel, 2008, Arias Chao et al., 2021) and Hard Drive Lifetime (Backblaze, 2022). In the case of TurboFan, NASA published a simulation dataset on turbojet engine failures, where each instance is labeled with the remaining useful life. On the other hand, Backblaze published statistics on different hard drive models, showing failure time. An interesting aspect of the latter is that it contains information from multiple HDD models, generating data compatibility problems between drive models.

Given the constant drift in real-world scenarios, we propose an additional condition the benchmark must meet: a dataset should have multiple distributions in the data. These multiple distributions can result from changes in the input or output distribution or even both. Ideally, this will allow training models to corroborate how they would behave in case of drift in the current data, verifying the generalization capability of the presented solutions. Once we have these benchmarks, PdM can expand to different scenarios and contexts, incorporating the idea of continually training a model to adapt to changing distributions.

As we will see in the following sections, non-stationary environments study the effect of distribution drift (Ditzler et al., 2015). By having data sequences that can change their distribution over time, we must train models that can either detect the change in the distributions or update their weights continually, avoiding affecting previous knowledge. Despite being an open research area, these techniques can contribute greatly to the PdM community.

**3. Non-stationary environments**

A traditional ML model learns by optimizing its weights based on all available data. The data used in training follows a distribution that reflects the reality seen so far, a reality that we assume does not change over time. Based on this assumption, the generalization of the model can be limited by the distribution of data on which it was trained.

Unlike closed or simulation environments, where we can control all variables, raw data does not follow a stationary distribution (Hu et al., 2020). In real-world applications, we have a constant flow of information, whose distribution can change due to various external or internal factors. These changes in the data distribution cause issues in previously trained models, mainly because their weights are not prepared to face drift concerning the training data. This problem creates the need to update the model continually.

Drift in the distribution can be defined in multiple ways (Quiñonero-Candela et al., 2008, Hu et al., 2020). Let $P_t(x, y)$ be a distribution where input data $x$ with label $y$ can be sampled in time $t$. In stationary environments, a model is trained to assume that $P(x, y)$ remains fixed regardless of time $t$. Instead, we can observe multiple examples of how the distribution shifts. One example is changing the probability $P(y \mid x)$, where the label associated with an input $x$ can change over time. A second example is changing $P(x)$, where the input distribution changes by adding novel concepts to previously seen classes or adding new classes to the problem.

**Table 1**
Summary of the benchmark datasets for Predictive Maintenance task. TS: Time-Series; DSC: DataSet Characteristics.

| Dataset name | Publish data | DSC | F. Diagnosis Classif. | F. Prognosis Regress. | Sources |
| --- | --- | --- | --- | --- | --- |
| Bearing Dataset | 2007 | TS | ✓ | - | Lee et al. (2007), Lessmeier et al. (2016), Nectoux et al. (2012) |
| Battery Aging | 2007 | TS | ✓ | ✓ | Saha and Goebel (2007) |
| Microsoft PdM | 2020 | TS | ✓ | ✓ | arnab (2020) |
| Robot Execution Failures | 1999 | TS | ✓ | - | Lopes and Camarinha-Mato (1999) |
| Bosch Production Line Performance | 2016 | TS | ✓ | - | Bosch (2016) |
| Turbofan | 2008 | TS | - | ✓ | Saxena and Goebel (2008), Arias Chao et al. (2021) |
| Hard Drives Lifetime | 2013-2022 | Tabular | ✓ | ✓ | Backblaze (2022) |

PdM scenarios are not an exception to these problems, primarily because models are trained on simulated or limited data sets. As drifts in data distribution occur, newly acquired data can differ from the trained model, decreasing the performance of the model. Internal factors, such as changes in sensors and components, and external factors, such as temperature variations, may lead to these drifts. Moreover, these problems can be aggravated in PdM, where the dependency between sensors can strongly influence fault detection (Alippi, 2014).

Due to the increasing popularity of non-stationary data applications and the increasing interest in PdM scenarios, it is necessary to identify the different types of drift and how they affect various models. A model trained under the false stationarity assumption will inevitably become obsolete, performing sub-optimally at best or failing catastrophically at worst (Ditzler et al., 2015). In Section 3.1, we will first define different types of distribution drift. Then, in Section 3.2, we will present several approaches to drift detection.

*3.1. Drifts types*

Just like in nature, data streams can have multiple distributions and variations. In the literature, different types of distribution drifts can be detected (Quiñonero-Candela et al., 2008, Ditzler et al., 2015, Hu et al., 2020, Lesort et al., 2021). Here we briefly explain those who generate the greatest consensus.

Assuming a data generation model $p(y|x)P(x)$, Quiñonero-Candela et al. (2008) propose different types of distribution drift:

1. Simple Covariate Drift: is when the marginal distribution $p(x)$ changes without affecting the posterior probability of classes. For example, the input condition changes for some assets due to internal or external factors.
2. Prior Probability Drift: is when the probability of $P(y)$ changes over time. For example, the definition of fault changes, or the time to perform maintenance decreases, changing when maintenance is required.
3. Domain Drift: is when changes in the marginal distribution $p(x)$ affect the posterior probability of classes, in other words, the map function changes. An example is adding or changing machines to which the model is applied. These modifications change the definitions of the model inputs that affect the output definition.
4. Sample Selection Bias: is when the training set does not accurately represent the test set distribution. This drift can happen in PdM when most of the collected data are from specific machines, for example, machines in a particular place. This limited collection of samples can be conditioned by certain internal factors of a particular location, like temperature and humidity.

Since the different types of drift are not independent, it is possible to face combinations of them. Fig. 3 represents the different types of drift.

In addition to the types mentioned above, drift can have a few characteristics. The first is the graduality with which changes take place. These can be **abrupt** or **gradual**. In the first case, drastic changes occur from time $t$ to time $t + 1$, mainly due to a failure or drastic change in the machine. On the other hand, a gradual change occurs slowly over time, either due to environmental change or machine component aging. These characteristics can be summarized in Fig. 4.

Another characteristic is whether they are **permanent** or **transient**. In the first case, the changes are expected to last for an indefinite period into the future. These variations can occur when a piece of machinery is changed or placed in a different environment. On the other hand, transient changes are temporary and only last a short time before returning to the previous distribution. These changes in distributions can occur when failures in related components appear and can be later solved.

Some transient changes can also be categorized as **recurrent**. This can occur when the distribution drifts follow cyclical patterns, as they can occur with external patterns, such as the seasons of the year.

*3.2. Drift detection*

A possible starting point for generating solutions is identifying when these distribution changes occur, regardless of the type of distribution drift. This motivation is precisely the purpose of drift detection, where methods and metrics are used to determine if data distribution has changed. When such a phenomenon occurs, a new classifier can be created, or the model weights can be updated or reinitialized. This depends on the problem objectives and constraints.

Following the categorization proposal of Hu et al. (2020), Lu et al. (2019), we can divide drift detection methods into two categories. The first is known as a performance-based approach, where the performance of the arriving data is monitored, and a significant drop (or spike) signals a possible change in the distribution. The second group is data distribution-based approaches, which are based only on data distribution. Since our primary goal is to present the problems as a solution for the PdM scenario, we invite more curious readers to review works by Gemaque et al. (2020), Iwashita and Papa (2019) for more details.

*3.2.1. Performance-based approach*

A natural way to verify when there is a change in the distribution is to check how the performance of a trained model evolves. If the performance remains relatively stable, one can assume that the current distribution is similar to the one used in training. On the other hand, if the performance of the model changes drastically or a progressive decrease is seen, distribution drift may have occurred.

One of the first methods proposed for drift detection was FLORA (FLOating Rough Approximation) (Widmer & Kubat, 1996), which consists of a family of algorithms. The framework identifies three concepts that approximate positive, negative, and potential future samples and determines the performance of each group. The Drift Detection Method (DDM) (Gama et al., 2004) proposes to train a new classifier when it detects changes above a certain threshold. The old classifier will be used until the detected changes do not reach a second threshold, at which the old classifier will be discarded and replaced by the newly created classifier. Several variants of DDM have emerged, as described by Gama and Castillo (2006), Baena-Garcıa et al. (2006), Frias-Blanco et al. (2014), Liu et al. (2017b), Xu and Wang (2017).

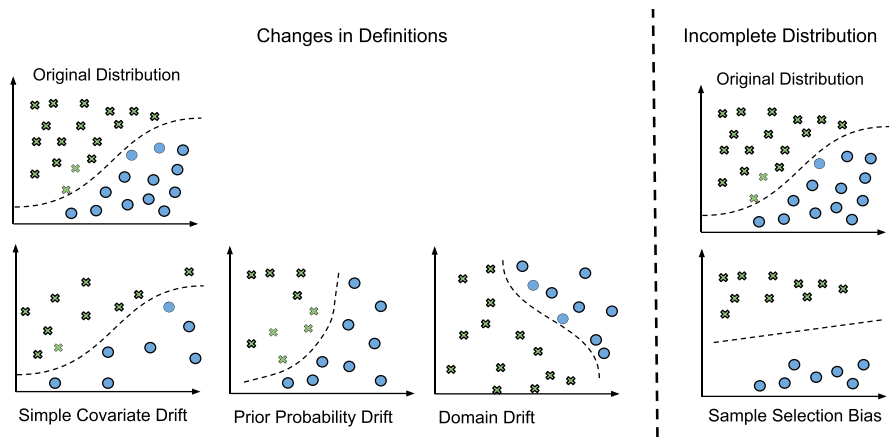

**Fig. 3.** Based on definitions in Quiñonero-Candela et al. (2008), we can see how the decision frontier changes when the available data distribution changes. We could train a flawless classifier if we could access a complete distribution that knows past and future definitions. However, this is impossible for two reasons: (1) The definitions we have of the current distribution can change, either by minor variations in the input data (Simple Covariate Drift), changing the actual data output (Prior Probability Drift), or changing the definition map function (Domain Drift). (2) Having a complete distribution of the input data is unlikely, which makes the model unable to find a suitable division. This problem can be caused by incomplete sampling on the current distribution (Sample Selection Bias).

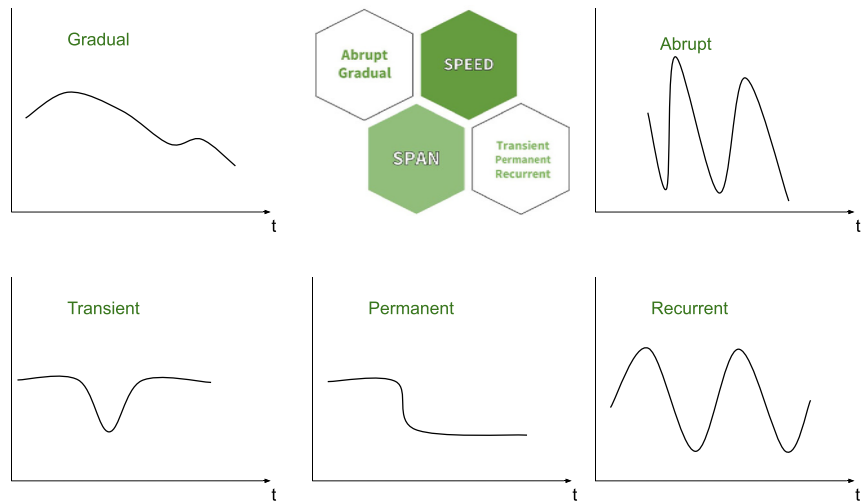

**Fig. 4.** Depending on how the drift changes over time we can divide it into two characteristics. The first characteristic is related to the speed of change. This can be gradual or abrupt over time. The second has to do with the span of time. It can be a transient, permanent, or recurring change.

Unlike DDM, which uses the error of the current batch to determine when a distribution drift occurs, ADaptive WINdowing (ADWIN) (Bifet & Gavalda, 2007) compares the performance of the current distribution with a window of previous distributions. Numerous methods have been developed from ADWIN, including Bifet and Gavalda (2009), Bifet et al. (2009a, 2009b), Gomes et al. (2017). A similar idea is proposed by Harel et al. (2014), where they employ random permutations in the data to analyze a difference in the loss to detect drifts.

Another alternative to the previous methods is what is proposed in Diversity for Dealing with Drifts (DDD) (Minku & Yao, 2011). In this work, the authors propose a combination of ensembles to generate diversity in performance to detect different types of distribution drifts. Several works have used ensembles to detect drifts, including those by Abdulsalam et al. (2010), Ortíz Díaz et al. (2015), Sethi and Kantardzic (2015, 2017).

*3.2.2. Data distribution-based approach*

Performance-based methods usually require labeled data to find drifts, so they cannot be applied to unlabeled or semi-labeled data. Consequently, approaches that quantify the dissimilarity between the current batch (window) and the memory have been proposed.

An intuitive way to solve this problem is using Information Theory. In the Information-Theoretic Approach (ITA) (Dasu et al., 2006), the authors propose to divide the data into bins using a KD-Tree. Then, one can calculate the density of each bin and compare it between data groups. An alternative to KD-Tree is Nearest Neighbor, as presented by Liu et al. (2018a). With these cluster-based and density-based techniques, hypothesis testing can be used to determine whether or not there is a drift distribution. Another alternative is to study and analyze how representations of a selected group change over time. Examples of this strategy can be found in Maletzke et al. (2017), Costa et al. (2018), Maletzke et al. (2018), Li et al. (2019a).

Other methods use principal component analysis (PCA) to reduce the dimensionality of the vectors and compare them in the latent space. For example, Liu et al. (2017c) examines the angle of the projected embedding with PCA to determine changes in the distribution. Trees can be used to split and simplify comparisons, such as Li et al. (2015), where a method defines cut-points to detect drifts. Several other works apply the idea of low dimensionality to detect drifts, including Song et al. (2007), Qahtan et al. (2015), Lu et al. (2016), Gu et al. (2016), Bu et al. (2016, 2017), Liu et al. (2017a).

A revolutionary generation of methods for Drift Detection has appeared with the arrival of Variational Auto-Encoders (VAE) (Kingma & Welling, 2013) and Conditional VAE (CVAE) (Sohn et al., 2015). Similar to traditional AutoEncoder methods, VAE aims to find an encoder capable of reducing the dimensionality of the input. However, in VAE, one encodes the input into a distribution in the latent space. The encoder is trained to return a mean and variance given the input, so detecting outliers (or changes in the distribution) is straightforward. These methods are used in multiple scenarios, such as social media sentiment analysis (Zhang et al., 2020c, Patil et al., 2021), detecting drift in videos (Suprem et al., 2020) and fault detection (Kim et al., 2023).

### 3.3. Closing remarks

The existence of these changes in the distribution requires the model to constantly adapt to changes. For the model to perform correctly, there are some essential characteristics:

1. Quick adaptation to changes: The model needs to learn the updated distribution efficiently, ideally with few labeled data.
2. Not forget previously learned distributions: The model should remember classes and past distributions, despite continual weight updates.
3. Know how to forget bias or outliers: The model inevitably learns biases or outliers from current distributions. However, it would be essential to identify which information or data is relevant to past and future distributions.

A variety of approaches will be discussed in the following section for continually training a model to take into account newly arrived distributions.

## 4. Continual learning

The constant drift of the input data causes the need to continuously adapt the trained models to perform correctly against these newly arrived distributions. Ideally, just as people learn, we want a model capable of accumulating additional knowledge. This is done by reusing and complementing what has been learned in the past with these novel distributions.

The idea of a lifelong learning agent has been proposed in several works in the past few years (Lu et al., 2019, Parisi et al., 2019). However, there is no categorical solution to the problem since proposals often come up against the plasticity-stability dilemma.

The dilemma can be defined as follows: as a model is trained, the structure is modified by optimizing its weights or adding additional nodes. These modifications are related to the plasticity of a model to learn new data. However, performance drops dramatically when we do not have access to previously collected data, so stability helps mitigate a phenomenon called catastrophic forgetting (McCloskey & Cohen, 1989). The dilemma lies in how much a model can change to learn new distributions without forgetting previous experiences.

The trade-offs due to the dilemma mentioned above can affect the solution used in a given scenario. In some contexts, we want the model to remember all past experiences by maintaining a high degree of stability. In some other scenarios, it is critical for the model to learn the new data distributions quickly, which requires a significant amount of plasticity. Generally, in the CL paradigm, it is critical to keep the data availability scheme in mind, since it often determines the methodology to use. As shown in Fig. 5, we can outline four different scenarios of data availability:

- Zero availability: at each timestep, the only data available is the current distribution. Past data cannot be saved or used in a subsequent timestep.
- Limited availability: data is available only in a given window of time. Only a limited number of experiences can be saved in memory. When another experience is acquired, it is added to the memory, and the oldest is discarded.
- Partial availability: for each experience, the method can only save a small percentage in memory. One can access the memory to remember previously learned distributions when training the newly acquired distribution.
- Full availability: the method has full access to past and current data. During the training process, the model has complete access to past distributions, not only to approximations of them.

Although it is tempting to think that training data is always available, this is not always the case, especially in real-world applications. There are multiple examples where this can occur, such as privacy concerns, limitations on the amount that can be saved, or simply because of data loss. Even if we have Full Availability, the cost of training the model increases considerably with each new experience.

In addition to data availability, several characteristics can affect how we deal with CL problems. Based on Iwashita and Papa (2019), Lu et al. (2019), we can divide the CL pipeline into different modules, which can be modified for each strategy. Here, data is continuously collected. These modules loop continuously to add new knowledge to the model, as shown in Fig. 6. Specifically, we can identify four modules in the pipeline:

1. Data collecting: defines which strategy will be used for storing and accumulating data from past experiences. These strategies can vary depending on the setting.
2. Data modeling: specifies how the feature vector is created. This vector depends on the model and data, but statistical methods, pre-trained models, or models trained from scratch can be used.
3. Drift detection: defines a method to recognize drifts, defining which action to take. For example, the method can trigger a model update when it detects a drift.
4. Model update: specifies the method for updating the model. Most research efforts in CL focus on this aspect.

We will center our discussion around the *model update* since it is responsible for defining how to adapt the model. We present two categories to update models when facing distribution drifts: First, in Section 4.1, we will discuss classical machine learning methods. The second category, discussed in Section 4.2, talks about some deep learning approaches, where we care about forgetting and reusing the information learned from previous representations and the classifier.

### 4.1. Adaptation methods

Detecting distribution changes can be considered the first step. However, how the model will react to these changes must also be defined. Traditionally, classical adaptation methods can be divided into two categories, depending on how the model reacts to distribution drift (Ditzler et al., 2015). The first category of approaches actively pursues distribution drift. These approaches first require detecting when a change in the distribution happens, then adapt the classifier to learn the new data distribution. The second category of approaches is known as passive approaches. Sometimes, detecting distribution drift is not possible, so instead of reacting when drift is detected, the model continuously adapts to the stream.

#### 4.1.1. Active methods

Active methods must first detect when distribution drift occurs before reacting to a change in distribution. Then, the classifier discards obsolete knowledge and adapts to the updated environment. It is critical to note that most methods discard previous knowledge by completely retraining previously learned models.

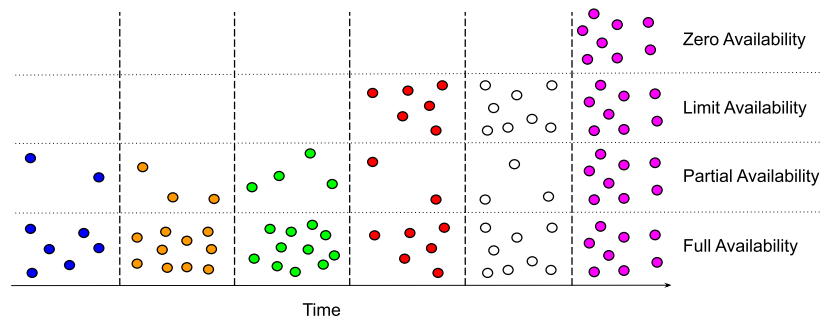

**Fig. 5.** Depending on the scenario and setup, different regimes can be found concerning data availability. For scenarios where there is no or limited data-saving option, we talk about zero or limited availability. On the other hand, if we can save a percentage or all of the data from each distribution, we speak of partial or full availability.

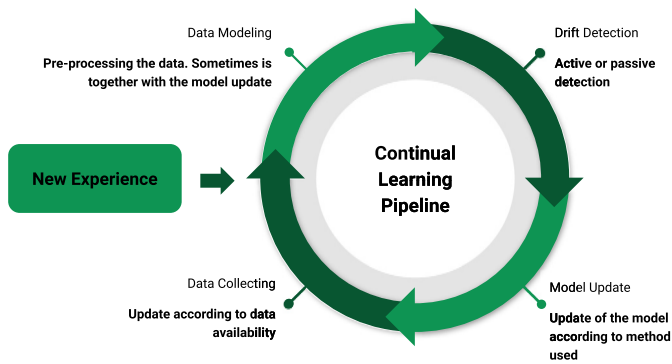

**Fig. 6.** We divide a continual learning solution into four modules. First, once an experience arrives, we need to pre-process the data and train the model. Then, if drift in the data distribution is detected, we update the model to face the updated data distribution according to the method we are using. Finally, once the experience is complete, we update our memory using the appropriate data collection technique.

According to Ditzler et al. (2015), active adaptation methods can be categorized as: *windowing*, *weighting*, and *random sampling*. In the first category, a sliding window saves a limited amount of the last samples. When drift is detected, one can discard the previously learned classifier and retrain another one using only samples from the window. This window can have a fixed user-defined length (Alippi & Roveri, 2008), or adaptive (Alippi et al., 2013, Bifet & Gavalda, 2006, Alippi et al., 2012, Cohen et al., 2008b, Alippi et al., 2011); in the latter, the size of the window changes depending on the distribution. This approach is usually associated with a *limited availability* of data.

The second category proposes methods that select a sub-group of elements used to retrain the models based on assigning weights according to either a relevance score or aging of the data. In Koychev (2000), Cohen and Strauss (2003), the authors propose different functions that gradually decrease the weights of previously learned samples, giving more importance to new examples. An alternative weight approach is presented by Alippi et al. (2009), Klinkenberg (2004), where a performance metric is used to determine the relevance of each example. The most significant disadvantage of these approaches is that they usually assume *full availability* of the data.

When data from previous experiences cannot be completely saved, one must look for alternatives. The third category proposes sampling data from experience to retrained classifiers. One of the most popular methods to select which elements to save is reservoir sampling (Vitter, 1985), in which a data point has probability $p$ of being inserted into memory. When the memory is full, the newly picked element will replace a random one already present in the buffer. Examples of how to use it in data streams can be seen in Aggarwal (2006), Ng and Dash (2008).

*4.1.2. Passive methods*

An alternative strategy to active methods is to assume a constantly evolving stream of information, resulting in models that are continuously updating. This solution allows independence concerning the correct and early detection of the distribution drift. However, these approaches can be ineffective in abrupt changes since they usually have a slower adaptation rate.

These methods can be divided into two strategies, according to Ditzler et al. (2015). The first strategy uses a single classifier model that is updated when new data arrives. One of the most used models in this context are decision trees (Domingos & Hulten, 2000, Hulten et al., 2001, Liu et al., 2009), where different metrics are applied to determine when to split a node. However, fuzzy-logic-based approaches (Cohen et al., 2008a) and neural networks (Ye et al., 2013) have also been employed.

The second strategy is to have an ensemble of classifiers. These ensembles can easily add new information by adding another classifier to the ensemble. At the same time, it can eliminate irrelevant information by removing classifiers with poor performance. These advantages are related to the *stability-plasticity* dilemma. A direct approach to ensembles is to train a new model each time another batch arrives (Tsymbal et al., 2008, Street & Kim, 2001, Scholz & Klinkenberg, 2005), and then gradually eliminate the classifiers with the poorest performance. Other methods propose adjusting the ensemble size depending on the context (Kolter & Maloof, 2007).

*4.2. Deep representation learning*

Deep learning techniques have been widely applied to learning representations from data, including classification and regression problems. By training both the representations and the classifiers, these methods usually produce better results. However, an additional source of forgetting appears when learning the representation. This creates an incentive to focus on these techniques and propose new methods to mitigate forgetting. These methods can be divided into three large groups: regularization-based, architecture-based, and memory-based. Similar to previous sections, the objective is not to propose an exhaustive compilation of previous works but rather to briefly discuss common approaches. More comprehensive reviews were made by: Parisi et al. (2019), Delange et al. (2021), Masana et al. (2020a), Qu et al. (2021).

*4.2.1. Regularization-based methods*

Since the modifications of the weights are primarily responsible for forgetting, limiting their updates would address the problem. However, these modifications are also responsible for learning upcoming tasks, meaning that the model needs some plasticity to adapt to newly arrived data. Regularization-based methods aim to solve this trade-off between stability and plasticity with regularization over the objective function. These approaches aim to balance the model so that it can learn new tasks while penalizing modifications to the relevant weights.

One way to regularize the model is to penalize the modification of weights relevant to previous tasks. Various techniques have been proposed to measure *relevance*. Some measures used are: the fisher matrix (Kirkpatrick et al., 2017, Lee et al., 2020), the gradient (Zenke et al., 2017, Aljundi et al., 2018), the uncertainty (Ebrahimi et al., 2019), among others (Aljundi et al., 2019c, Saha & Roy, 2021). Alternatives to decrease interference between tasks are presented in Lomonaco et al. (2020), Masana et al. (2020b), Hurtado et al. (2021a). The authors propose freezing previously trained weights to eliminate interference while inhibiting information transfer between tasks. Araujo et al. (2022b) proposes to dynamically regularize each layer depending on the entropy of the representations.

Another alternative is to regularize at the level of the representations found. Based on knowledge distillation, these methods aim to have the representations of previous tasks change as minimally as possible. The precursor of this idea is Learning without Forgetting (LwF) (Li & Hoiem, 2017), in which the classification of new data in an old classification head is optimized, following the idea of soft targets (Hinton et al., 2015). Based on the idea of LwF, different modifications have been proposed to overcome further limitations (Jung et al., 2016, Rannen et al., 2017, Dhar et al., 2019, Zhang et al., 2020a). Additionally, Hou et al. (2018), Lee et al. (2019) proposes distilling the information of an old task into a more effective model (student-teacher approach) using external data.

These approaches generally perform well on problems involving fewer tasks. However, problems such as accumulated drift in weight values and interference make these approaches challenging to scale when the number of tasks increases.

*4.2.2. Architecture-based methods*

Another alternative to mitigate forgetting is to force each task to use specific weights. Instead of interfering with previously acquired knowledge, the model learns patterns specific to the current task. By adding new weights, the model maintained its plasticity and mitigated forgetting to some extent.

Works from Rusu et al. (2016), Fernando et al. (2017) propose cloning a model and adding connections between the models, creating an information exchange from old to new tasks. Rather than cloning or modifying the base model, some works propose adding extensions, modifying the output without interfering with the base knowledge (Serra et al., 2018, Ebrahimi et al., 2020). As a drawback, the amount of disk space required by the model increases linearly with the number of tasks. A popular approach to overcome this problem is to incorporate trainable binary masks to select parameters either through pruning (over a backbone model) (Mallya & Lazebnik, 2018, Mallya et al., 2018), or using the Lottery Tickets Hypothesis (Wortsman et al., 2020).

The main problem with these approaches is that the model tends to grow linearly with the number of tasks. This problem is partly mitigated by methods that add only sub-parts to the model (Ebrahimi et al., 2020, Hurtado et al., 2021b) or seek to reuse knowledge (von Oswald et al., 2020, Mendez & Eaton, 2021), but it remains a problem when space is limited. Another critical problem, especially in real-world applications, is knowing the corresponding task of each example in training and inference. Since each example needs to identify the corresponding weights to be used to predict.

*4.2.3. Memory-based methods*

As mentioned, forgetting is caused by a change in weight caused by a drift in the training distribution. A subset of elements from previous distributions can be added to the training set of the new experiences to mitigate the drift. With this approach, the model can learn from the new distribution while retaining information from the past. Solutions can follow two main alternatives: 1) save raw examples (Hu et al., 2019, Riemer et al., 2019, Rebuffi et al., 2017, Ebrahimi et al., 2021, Lopez-Paz & Ranzato, 2017, Chaudhry et al., 2019a) (or features Iscen et al., 2020, Caccia et al., 2020) of past experiences, and 2) generate examples (Lesort et al., 2019, Shin et al., 2017) (or features Kemker & Kanan, 2018, Xiang et al., 2019, Hayes et al., 2020) from previous tasks.

One of the most popular works that use raw examples is iCaRL (Rebuffi et al., 2017), in which elements are stored based on the distance toward a class prototype. Without using data from previous tasks directly in training, works from Lopez-Paz and Ranzato (2017), Chaudhry et al. (2019a), Aljundi et al. (2019b) seek to minimize the interference between stored data and the current task. Due to the high cost of saving samples from previous tasks, Iscen et al. (2020), Caccia et al. (2020) propose to save feature vectors. This solution reduces privacy and memory concerns, as vectors typically require less memory than complete elements.

In case it is not possible to save real examples, an alternative is to train generative models that approximate previous tasks. Several methods use GANs to generate examples of past tasks (Shin et al., 2017, Kemker & Kanan, 2018), but different generative models have been studied (Lesort et al., 2019). An alternative is presented by Hayes et al. (2020), where it saves compressed representations using product quantization instead of training generative models.

Memory-based methods require many examples to be saved for the correct model behavior. Mainly because these examples should accurately represent the distribution of previous tasks. Other limitations are privacy or storage restrictions.

Due to such limitations and restrictions, researchers have been working on approaches that emulate memory-based approaches, without explicitly saving patterns in a buffer. In this regard, a promising research direction is prompting (Liu et al., 2021). Prompting is used to enrich the model input with additional information to obtain better predictions.

Prompting technique has recently been studied in CL in Learning to Prompt in Continual Learning (L2P) (Wang et al., 2022c). Here, the prompts are small, learnable parameters saved in memory. These prompt vectors are concatenated with the input pattern to produce enhanced representations. It has been shown that catastrophic forgetting can be greatly mitigated by enriching input patterns while using less memory. In DualPrompt (Wang et al., 2022b), an extension was proposed, where the authors demonstrate how prompts should be distinguished between task-specific and task-invariant.

An essential characteristic of memory-based methods is how to populate the memory. A simple but efficient method is the reservoir strategy (Vitter, 1985), which randomly selects elements that are put into memory. Some strategies populate the memory based on the representativeness of the samples, taking into account different metrics (Chaudhry et al., 2019b, Hayes et al., 2020, Hayes & Kanan, 2021, Aljundi et al., 2019a). However, despite the popularity of these methods, more needs to be studied about the composition of the memory (Tiwari et al., 2022, Araujo et al., 2022a).

## 5. Real-world applications of continual learning

Although most CL works are presented in laboratory contexts, some focus on transferring knowledge to real scenarios. There are various practical fields where researchers and companies have applied CL. Publications in computer vision (Alzubaidi et al., 2021), Machine Learning Model Operationalization Management (MLOps) (Hewage & Meedeniya, 2022), robotics and machine vision, industry, and edge computing, have expanded knowledge from the theoretical to more practical situations. An overview of some of the latest and most relevant CL studies for different applications is presented in the following sections:

## 5.1. Continual learning for vision and embedded devices

As presented in Section 3, real-world applications do not follow the rules of traditional machine learning algorithms, where it is assumed there is a static training set with independent and identically distributed (iid) data. Most real-world learning problems involve dynamic and non-stationary environments, and many are applied to embedded devices to minimize latency. Moreover, continuous on-device learning is critical to preserve user privacy and security.

In Hayes and Kanan (2022), Pellegrini et al. (2020), CL is studied in embedded devices. The first paper examines CNN performance in different CL environments, in terms of memory requirements and computational tradeoffs. In the second paper, the authors implement CL strategies for Android smartphones. A "Latent Replay" approach is proposed, in which intermediate representations are stored instead of raw data. To keep the representations valid, the first layers of the model are slowly updated, and only the last layers are free to learn at full pace. These results provide evidence that state-of-the-art CL approaches, extended with latent replay, can learn efficiently and provide a positive computation-storage-accuracy trade-off.

## 5.2. Continual learning enables in MLOps and model service computing

Another practical challenge to applying ML in real-world applications is the framework and pipeline used to offer and update the model as a service. Recent works try to overcome practical challenges in cloud computing and ML pipeline using CL. In contrast to the classical complementary algorithmic perspective, Diethe et al. (2019) describes infrastructure and systems engineering from a different perspective. Here, the author proposes Auto-Adaptive ML systems, an architecture for self-maintaining, which relies on continual learning to manage ML models in production effectively. Each system component deals with adaptation to shifting data distributions, coping with outliers, retraining when necessary, and adapting to novel tasks.

A second proposal is presented by Semola et al. (2022), which combines CL infrastructure with algorithms. It defines a novel paradigm called Continual Learning-as-a-Service (CLaaS), which combines continual machine learning and continuous integration techniques. In the ML pipeline, CLaaS can be used to support fast industrial R&D prototype projects based on the MLOps process and Continuous Monitoring and Training. Thus, ML models can be trained and updated in a serving system efficiently, scalable, and adaptable. It enables the creation and maintenance of low-cost smart infrastructure without requiring a deep understanding of CL. Results on real-world cases, such as robotics, object recognition, and the fashion domain, show the efficiency and usability of CLaaS in continual Cloud-Edge training.

## 5.3. Research direction in continual learning for industries

A major problem when applying CL in natural settings is the disjoint between lab-context scenarios and real-world scenarios. Current CL methods assume a disjoint set of classes per experience, which is not always realistic in real-world applications. Bang et al. (2021) proposes a promising research direction, in which the authors focus on the importance of diversity of samples in episodic memory when training with not-disjoint classes per task. The authors call this setting Blurry CL, which is presented together with a novel memory management strategy called Rainbow memory. This novel strategy selects samples based on classification uncertainty and data augmentation.

Due to the limitations of Blurry CL, Koh et al. (2022) proposes a more realistic and practical CL setting that adheres strictly to an online and task-free setup. Here the authors propose an innovative CL setup called i-Blurry, which takes advantage of blurry and disjointed settings. While it assumes that the model continues to encounter new classes as tasks arrive, it also assumes that classes overlap between tasks. This setup is similar to what we can find in real applications, where new or old classes are encountered together. In the same paper, the authors propose a novel metric to better measure a CL model's capability for the desirable any-time inference called *area under the curve of accuracy* ($A_{AUC}$).

One disadvantage of i-Blurry is that it assumes clean labels during data streams, which is not always feasible in real-world applications. Due to this, an improvement is presented by Bang et al. (2022), where a similar scenario is presented, but corrupt labels may appear in the stream. This work stresses the importance of diversity and purity of examples in the memory, by using a semi-supervised approach that uses label noise awareness, diverse sampling, and robust learning.

Recent studies have focused on addressing a more realistic CL setup, but there is still much to be done. The need for a practical CL setting also seems to be a starting point for businesses to focus on CL. Similar to PdM benchmarks, these new setups must meet real applications' standards to be a valid point of comparison between methods.

## 6. Continual learning for predictive maintenance

Despite the effort to apply CL to various real-world applications, there are still some challenges and limitations. Because of this, in this section, we will discuss the current situation in applying CL specifically to PdM. Then, considering the current state-of-the-art and what has been discussed throughout the document, we will expose the present challenges and promising future research directions.

### 6.1. Previous works on CL for PdM

Some works have followed the trend toward continually adapting deep neural network models for real-world applications. Specifically in PdM, Maschler et al. (2020) examined the effect of Elastic Weight Consolidation (EWC) (Kirkpatrick et al., 2017) in the TurboFan benchmark. Because of legal and technical reasons, it is not easy to accumulate large amounts of data, making it infeasible to centralize vast amounts of samples in real-world applications. Due to this problem, Maschler et al. (2020) proposes a CL setup to facilitate training on small and decentralized datasets (embedded devices) over time. The results show that EWC can improve the model when sub-datasets are not independent, increasing knowledge transfer between the different sets and reducing forgetting.

A second paper focused on PdM is by Maschler et al. (2021). Here the authors propose an extension to Maschler et al. (2020) by comparing different regularization-based methods in the lithium-ion battery benchmark. Similar to the previous work, the benchmark is adapted so that each experience is a set of different batteries. This adaptation simulates a situation where we want to transfer knowledge from one battery model to another. Using an LSTM as a base architecture and EWC, Online EWC, and Synaptic Intelligence (SI) as CL methods, the authors show how regularization methods help to forget less and improve knowledge transfer between tasks.

In a third work, Chen et al. (2022) based the solution on iCaRL and LUCIR. They use dual residual blocks to address the plasticity-stability dilemma. The first block is a steady learner that preserves knowledge by updating its weights slowly. On the other hand, the second block can quickly adapt to newly acquired knowledge. Residual blocks are integrated into an adaptive branch that adapts weights to balance stability and plasticity. A distillation loss is added to mitigate forgetting at a feature level.

Another approach that uses CL in fault detectors is proposed in Gori et al. (2022), where the authors combine RNN and Kullback-Leibler divergence with CL for anomaly detection on turbomachinery prototypes. Here, using limited data availability, the authors continually retrain an LSTM model to adapt to the data distribution drift.

An area related to PdM is quality prediction (Kashyap & Datta, 2015). A work that applies CL to this problem is Tercan et al. (2022). Here, the authors use the injection molding problem to generate a synthetic dataset with 16 different structures, which are trained sequentially. Using an MLP and Memory-Aware Synapse (MAS) (Aljundi et al., 2018) as a regularizer, the authors conclude that applying CL can help reduce forgetting of previously learned tasks and increase knowledge transfer when there is limited training data.

Recently, a survey about transfer learning for fault diagnosis addressed some of our concerns (Li et al., 2022). However, in most of those approaches, one must assume access to the source and target domains. This is an assumption that is not always feasible in CL scenarios, as discussed in previous sections.

## 6.2. Deep learning challenge in predictive maintenance

The success of data-driven methods in different applications is palpable. Deep learning is used in most PdM solutions, as shown in Section 2. However, as we have discussed, the dynamic nature of predictive maintenance brings challenges to static machine-learning solutions that are far from being solved.

As exposed by Escobar et al. (2021), Tercan et al. (2022), the same machine can behave differently in different contexts or environments, producing noise in the prediction of when maintenance is necessary. The lack of adaptation is that most current models are trained only with static and limited or synthetic data distributions, leading to a lack of generalization. We expect that a similar issue will arise with small technical differences in embedded devices. This is where extrapolating the knowledge from one device to another without a large amount of training data is normally a challenge. Hence, developing methods that enable continual adaptation to different environments is crucial.

A popular solution to continual adaptation is training with the whole sequence of data. However, whether for legal reasons or company criteria, not all data is maintained (Maschler et al., 2020). This creates the need to train with a small data distribution from scratch or use a pre-trained model. Several works have shown that it is better to learn from a trained model than to train one from scratch when having a reduced amount of data (Maschler et al., 2020, 2021, Tercan et al., 2022). For this, transfer learning seems like a plausible solution. However, this technique does not solve the problem of forgetting previous distributions. CL aims to develop methods that avoid forgetting by taking advantage of prior knowledge.

Another limitation of most pre-trained models is their training complexity. The constant stream of newly acquired samples makes it challenging to adapt big models with only a small batch of data. It would be helpful if more methods were proposed with tools for making small and rapid adjustments to achieve good performance in the current task and avoid drastic changes in the weights.

The dynamic nature of PdM has made reinforcement learning (RL) methods a viable option to solve the problem since they can interact with the environment through actions that produce a result and provide feedback. In general, RLs solutions consist of a feedback process loop where, given a state, an agent interacts with an environment by carrying out an action that can provokes changes in the state (Siraskar et al., 2023). A PdM problem can be proposed as a Markov Decision Process (MDP) and the reward function is based on the productive up time (Zhang et al., 2019a, Ong et al., 2021, Siraskar et al., 2023). However, similar to deep learning methods, these techniques are not exempt from problems of drift in the distribution (Khetarpal et al., 2022, Moos et al., 2022), since it learns to make decisions based on an environment where external factors can influence their actions.

In order to obtain robust solutions and perform well on the discussed challenges, it is imperative to have a robust benchmark that allows us to evaluate our approaches efficiently. These benchmarks must reflect the dynamism we seek in non-static environments.

It is pertinent to note that many real-world applications are transitioning through the same process, from stateful to dynamic environments, as presented in Section 5. As a result, CL can enable effective learning over time on smaller, decentralized datasets without centralized server storage. It is crucial to transfer knowledge from other areas to the challenges presented in this section.

## 6.3. Research directions for CL in PdM

Literature on CL strategies for PdM domains is still emerging. However, we believe it is essential that the research community trends towards methods in non-stationary environments. Regarding current advances and research topics, we present a non-exhaustive list of possible topics to contribute to the future of CL for PdM.

**Exploring novel methods:** As of now, few PdM works have used CL strategies to mitigate forgetting, focusing only on regularization-based methods. However, replay strategies are significantly more effective at mitigating forgetting, even if they require a small amount of previously learned data. Further research on this and architecture-based methods is needed to validate different scenarios and contexts.

**Sequential data:** As most PdM problems are based on sequential data, it seems essential to focus on applying CL in recurrent networks. There has been a growing interest in applying recurrent neural networks in continuous learning settings (Cossu et al., 2021). Despite the advances, it is still an open research problem that lacks a proper benchmark and could benefit the PdM area.

**On the edge prediction:** In many real-world applications, data is collected on-the-edge devices. These devices are often too small to train or modify large models. This is even more complex if we consider the speed at which newly collected data can be obtained. In this sense, research methods must adapt to a novel task quickly, making small but essential updates without forgetting the base knowledge. Moreover, it is crucial to find frameworks that facilitate communication and knowledge transfer between these devices. This research direction is particularly crucial for PdM, where different sensors are used to collect information.

**Non-stationary environments:** It is not always the case that we need to consolidate acquired knowledge without forgetting previous tasks. The idea of continually accumulating knowledge can improve performance on future tasks due to transfer learning but it also creates a more complex scenario. There may be situations where it is enough to constantly adapt the model to the evolving distribution without worrying about previous knowledge. Given the limited data available, it is vital to propose methods that quickly adapt to various distributions without overfitting.

**Realistic benchmarks:** As stated in the paper, current CL benchmarks do not reflect natural environments' complexity, making it difficult to verify the effectiveness of different methods in real environments. On the other hand, most PdM benchmarks do not consider the non-stationary component of the environment. Works like Bang et al. (2021), Koh et al. (2022), Bang et al. (2022) help point in the right direction, but a benchmark appropriate for PdM problems has yet to be developed.

**Pre-trained strategies:** Pre-trained models can be used as a knowledge base where methods extract information when needed. However, which information to extract and how to add relevant knowledge to pre-trained models is still an open question. Ideally, we need methods that require few adaptations to select this information. In addition to being more efficient, this also reduces the risk of forgetting since fewer weight changes are made. This topic also raises the question of the strategy used to train these models. A supervised strategy is usually used, but self-supervised strategies can prove to be useful in training weights to mitigate forgetting (Shi et al., 2022).

**Transfer learning:** Since multiple models can run in parallel on various edge devices, it is essential to develop a framework to improve communication and transferability of knowledge between models. It is essential to train models capable of correctly transferring what they have learned and incorporating knowledge from other devices. This is without affecting the performance of individual devices.

**Digital twins technology:** Another area that has grown considerably is digital twins (Barricelli et al., 2019), which seeks to create virtual replicas of object models based on real-time information and past experiences. Using the constant flow of data, digital twins can predict when a machine needs maintenance. Similar to PdM, this stationary environment can cause problems in simulation or prediction when external factors such as temperature, oscillations, or other components change. For these approaches to generalize to unknown distributions, they need to constantly adapt to evolving environments (Wang et al., 2022a). For this reason, it is necessary to apply and complement CL techniques to both adapt digital twins to non-stationary environments and continually train predictive maintenance methods.

## 7. A new structure for future benchmarks

As discussed in Section 5, some effort has been invested in applying CL in natural environments and bridging the gap between lab and real-world scenarios. However, a more comprehensive dataset and a more realistic sequence of tasks are still required. This section presents some limitations of current CL benchmarks and a novel proposal for creating sequences closer to natural environments.

### 7.1. Definition and limitations

Following the mathematical definitions by De Lange et al. (2021), the goal of CL is to train a sequence (possibly infinite) of tasks (or experiences) continually. Each task $t$ is represented as a dataset $D(x^{(t)}, y^{(t)})$ sample from a distribution $D^{(t)}$, were $x^{(t)}$ is the set of samples for task $t$, and $y^{(t)}$ the corresponding label. Since it is expected to train the model continually, methods minimize the following:

$$\sum_{t=1}^{T} \mathbb{E}_{x^{(t)}, y^{(t)}}[\ell(f_t(x^{(t)}; \theta), y^{(t)})] \qquad (1)$$

with limited or no access to previous data $(x^{(t')}, y^{(t')})$ when training tasks $t > t'$. CL methods seek to optimize the parameters $\theta$ by minimizing the loss expectation for all tasks in the sequence $T$.

Despite not being represented in the current definition, for simplicity and reproducibility, most CL benchmarks assume two characteristics when creating a stream of data:

1. Balance set of tasks: Each task is composed of a similar amount of elements and/or classes. For task $|D^{(t)}| \approx |D^{(t')}|$ and $|C \in D^{(t)}| \approx |C \in D^{(t')}|, t \neq t'$
2. Disjoint set of tasks: Each task is composed of a set of classes that do not appear in other tasks. Meaning that given a sample $p \in C_i$ and a class $C_i \in D^{(t)}, \neg \exists q \in C_i \wedge q \in D^{(t')}, t \neq t'$.

Due to previous assumptions, current CL benchmarks create unrealistic streams where each task has a similar amount of classes and data. In real-world applications, it is common to have a large amount of data to train (or fine-tune) the model in the first task, and smaller subsequent tasks that contain novel samples from new or old classes. Adding supplementary knowledge to the model is intrinsic to CL methods; however, in most previous benchmarks, proposals focused on only introducing novel classes, assuming that earlier tasks had access to the complete class $c$ distribution. In most scenarios, this assumption is incorrect. Therefore, it needs to be clarified how current methods behave when old classes need to be updated together with adding new classes.

### 7.2. Mitigating assumptions

To tackle the first assumption, and following how real applications' data distribution behaves, an easy-to-apply solution is to assume a large amount of data to train the first task. A $D^{(1)}$ that includes multiple classes can help the model train weights that generalize to unknown classes. This idea follows the principle of pre-training models and transfer learning, where it is common to train a model with a large amount of data, so it serves as a suitable starting point for upcoming tasks.

The previous idea can also be applied to unlabeled data. Using a large data set, $t = 0$ (pre-training), we can train a model with unsupervised/self-supervised techniques to obtain robust initialization for future tasks. This strategy constitutes a reliable starting point for multiple tasks and can even super-pass supervised initialization (Radford et al., 2021). It has also been shown that pre-training can mitigate forgetting by using a technique known as Continual Pre-training (Cossu et al., 2022b), which can add new knowledge and adapt the model without drastic modification in the weights.

The second assumption is more challenging to tackle. There are works where elements of the same class have been separated into different tasks. Lomonaco and Maltoni (2017) propose the scenario *New Instances* (NI), where novel instances of known classes can be presented in future tasks. This scenario allows the model to update knowledge from previously learned classes without adding classes. In addition, the authors propose the scenario *New Instances and Classes* (NIC), where upcoming instances of new and old classes appear in upcoming tasks.

The problem is that to create the sequence $T$, NI and NIC randomly sample elements from $D^{(t)} \sim D$. This selection means that for each class $c \in t$, the data follows $D_c^{(t)} \sim D_c$. Given that a similar distribution of class $c$ is learned in different tasks, only minimal or no knowledge is added to the model. These instances are only used as replays to remember past distributions. Fig. 7a shows how 3 tasks follow a similar distribution to the original dataset.

### 7.3. Concept-based splitting for realistic benchmarks in CL and PdM

Here we propose to extend previous work and split each class into concepts, dividing the concepts into different tasks. Each concept $k$ will follow a subset of the $D_c$ distribution. Each of these subsets represents a sub-class of $c$, dividing the class into groups that share a common factor (the class) but with significant differences. This proposal follows the idea of Sample Selection Bias presented in Section 3.1, where training a model with a subset $D^{(t)}$ of the complete distribution $D$ is unlikely to generate a model that fits the complete distribution. By forcing $D_c^{(t)}$ to adopt a different distribution than $D_c^{(t')}$, where $t \neq t'$, we expect the model to learn to adapt to a distinct distribution inside the same class.

This split of concepts inside a class helps understand how a model behaves when it must adapt to new classes and changes in known class distributions simultaneously. CL is designed to adapt to novel distributions continuously, and adding this enriched dimension can extend the applications where CL can be applied. Current CL scenarios only change the distribution by adding new classes or changing the input. However, it has yet to be thoroughly evaluated against more challenging benchmarks.

In addition to its advantages for classical CL problems, this scenario is closer to PdM problems. In PdM, the number of classes varies slightly, and the distribution of data constantly changes. For example, the Hard Drive benchmark Backblaze (2022) constantly adds new and old HD models with different capacities to the dataset. Even though identical HD with different capacities share common knowledge, it can change their failure rate (change in the distribution within the same class). Each capacity represents a concept $k$ where the classes $c$ are represented by the HD model, and the distribution of $D_c^{(t)}$ can differ from $D_c^{(t')}$ since the model can modify its features. Similar to the previous example, more benchmarks can be created following the ideas presented in this section.

## 8. Conclusions

Predictive Maintenance (PdM) is an area that has grown considerably recently, mainly due to the favorable results obtained by deep learning methods. Multiple efforts have been made to apply these methods to different problems, areas, and services. Despite the promising results, not much effort has been put into understanding the limitations of these methods in real-world environments for the PdM problem.

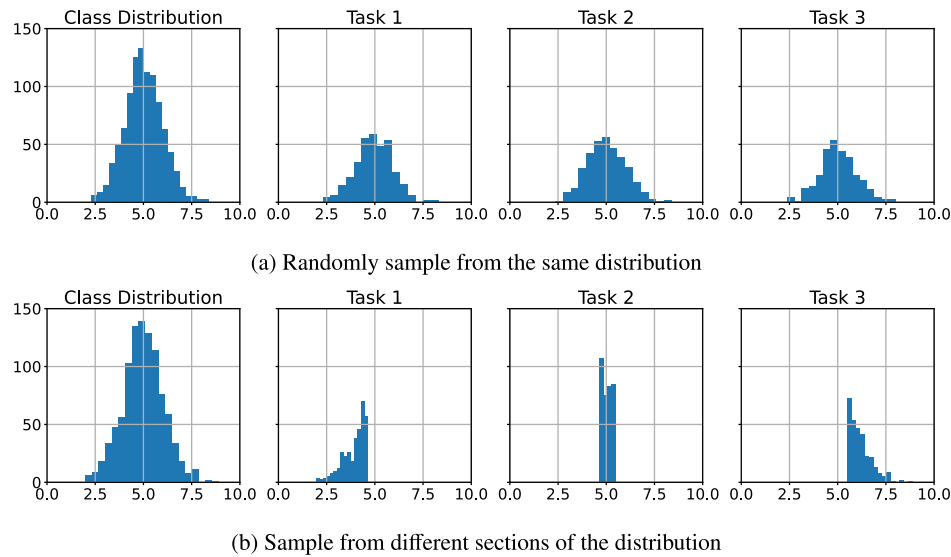

**Fig. 7.** A class distribution (represented by the figures on the left) can be split in different ways to shape the tasks or experiences. The upper part shows how random division creates similar distributions for each task. This means that there is not a large drift in the stream. Instead, the figures at the bottom show a division by concepts, where a clear difference is seen in the distributions formed in each task.

In this paper, we raise concerns about using deep learning techniques in PdM problems without considering their limitations. Here we focus on the idea of non-stationary environments and how this can affect deep learning models in PdM scenarios. We began by exploring PdM and detailing current benchmark limitations. Next, we introduced non-stationary environments and Continual Learning (CL) and then discussed current work and future directions at the intersection of the areas. Rather than ignoring these limitations, we emphasize that they should be addressed before they adversely affect machines and human lives.

With examples, we demonstrate that PdM scenarios are not exempt from these problems. At the same time, we present the area of CL as a possible solution, but with the problem that it does not have adequate benchmarks for the PdM problem. Aiming to contribute to the integration of both areas, Section 7 presents in detail what we believe are the limitations of current CL benchmarks and why the use of these scenarios in real problems is not direct. The last section proposes a more realistic structure for creating benchmarks that reflect real-world applications, which could improve the applicability of current methods and open up new research directions.

Along with being a starting point for discussing the limitations of current benchmarks used in CL applications, this paper presents an introduction to techniques and scenarios of CL for those working in PdM. In addition, it will be an introduction to real-world applications and scenarios for the CL community, which has always strived to find a direct application. Additionally, we present some limitations and future directions on the intersection between PdM scenarios and CL methods. Emphasizing one of the research directions, we discuss in detail the limitations of the current CL benchmarks. As a final contribution, we also explain why it is necessary to rethink how these benchmarks are generated.

The importance of creating an improved benchmark is crucial for PdM and CL. On the one hand, it would help the PdM area to extend to non-stationary environments, addressing problems such as the lack of generalization or predicting elements outside the training distribution more easily. Furthermore, it allows CL to expand beyond purely theoretical or toy benchmarks for more practical applications. As an example, a model trained with data collected from a particular type of wind turbine may fail if we apply it to another type of turbine, or if weather conditions change. Currently, to overcome this problem it is necessary to collect original data and train the model from scratch. CL techniques can change the pipeline by adapting the current model to a different context with a smaller amount of data and without having to learn from scratch, aided by forward transfer.

Addressing these limitations can drive the PdM area to new problems, expanding its applicability. Furthermore, the CL area would benefit from a different approach and more realistic solutions.

**Declaration of competing interest**

The authors declare the following financial interests/personal relationships which may be considered as potential competing interests:

Julio Hurtado reports financial support was provided by Sea Vision S.r.l.

**Data availability**

No data was used for the research described in the article.

**Acknowledgements**

This document is the results of the research project funded by SeaVision s.r.l., number 489999.